\documentclass[10pt,twocolumn,letterpaper]{article}

 \usepackage{iccv}              % To produce the CAMERA-READY version

\usepackage{graphicx} % for pdf, bitmapped graphics files
\graphicspath{{figures/}}
\usepackage{amsmath} % assumes amsmath package installed
\usepackage{algorithm}
\usepackage{algorithmic}
\usepackage{xcolor}
\usepackage{subcaption}
\usepackage{array}

\usepackage{multirow}
\usepackage[hyphens]{url}
\usepackage{bm}

\definecolor{iccvblue}{rgb}{0.21,0.49,0.74}
\usepackage[pagebackref,breaklinks,colorlinks,allcolors=iccvblue]{hyperref}

\def\paperID{10259} % *** Enter the Paper ID here
\def\confName{ICCV}
\def\confYear{2025}

\title{GraphEnet: Event-driven Human Pose Estimation with a Graph Neural Network}

\author{Gaurvi Goyal\\
Maastricht University \& \\
Istituto Italiano di Tecnologia\\
{\tt\small gaurvi.goyal@} \\ {\tt\small maastrichtuniversity.nl}
\and
Pham Cong Thuong\\
Istituto Italiano di Tecnologia\\
{\tt\small cong.pham@iit.it}
\and
Arren Glover\\
Istituto Italiano di Tecnologia\\
{\tt\small arren.glover@iit.it}
\and
Masayoshi Mizuno\\
Sony Interactive Entertainment Inc.\\
{\tt\small masayoshi.mizuno@sony.com}
\and
Chiara Bartolozzi\\
Istituto Italiano di Tecnologia\\
{\tt\small chiara.bartolozzi@iit.it}
}

\begin{document}

\maketitle
%\thispagestyle{empty}
%\pagestyle{empty}

%%%%%%%%%%%%%%%%%%%%%%%%%%%%%%%%%%%%%%%%%%%%%%%%%%%%%%%%%%%%%%%%%%%%%%%%%%%%%%%%
\begin{abstract}    
%%%%%%%%%%%%%%%%%%%%%%%%%%%%%%%%%%%%%%%%%%%%%%%%%%%%%%%%%%%%%%%%%%%%%%%%%%%%%%%%
% \todo{[TODO]}
Human Pose Estimation is a crucial module in human-machine interaction applications and, especially since the rise in deep learning technology, robust methods are available to consumers using RGB cameras and commercial GPUs. On the other hand, event-based cameras have gained popularity in the vision research community for their low latency and low energy advantages that make them ideal for applications where those resources are constrained like portable electronics and mobile robots. In this work we propose a Graph Neural Network, GraphEnet, that leverages the sparse nature of event camera output, with an intermediate line based event representation, to estimate 2D Human Pose of a single person at a high frequency. The architecture incorporates a novel offset vector learning paradigm with confidence based pooling to estimate the human pose. This is the first work that applies Graph Neural Networks to event data for Human Pose Estimation. The code is open-source at \url{https://github.com/event-driven-robotics/GraphEnet-NeVi-ICCV2025}.
\end{abstract}
%%%%%%%%%%%%%%%%%%%%%%%%%%%%%%%%%%%%%%%%%%%%%%%%%%%%%%%%%%%%%%%%%%%%%%%%%%%%%%%%
\section{INTRODUCTION}
%%%%%%%%%%%%%%%%%%%%%%%%%%%%%%%%%%%%%%%%%%%%%%%%%%%%%%%%%%%%%%%%%%%%%%%%%%%%%%%%

Visual Human Pose Estimation (HPE) is the task of estimating major body joints of a human agent in a scene. It is a crucial module in systems involving human-machine interaction, with the location and motion of the body joints used by downstream tasks like human action recognition, motion detection, motion analysis, gesture recognition, emotion detection amongst others~\cite{zhang2021SingleHPESurvey,wang2021deep,liu2022recent,chen2020monocular}. The maximum action speed for accurate HPE is constrained by the camera in the system. Fast human motion (e.g. during sports) or fast camera motion (e.g. on a flying drone) induces motion blur in the image, reducing the information available to the algorithm to detect precise joint locations.

\begin{figure}
    \centering
        % \subcaptionbox{Latency vs Accuracy Trade-off}{\includegraphics[width=0.75\linewidth]{publications/ICCV2025_HPE_GNN/Images/performance_tradeoff_eH36M.png}}
       {\includegraphics[width=0.75\linewidth]{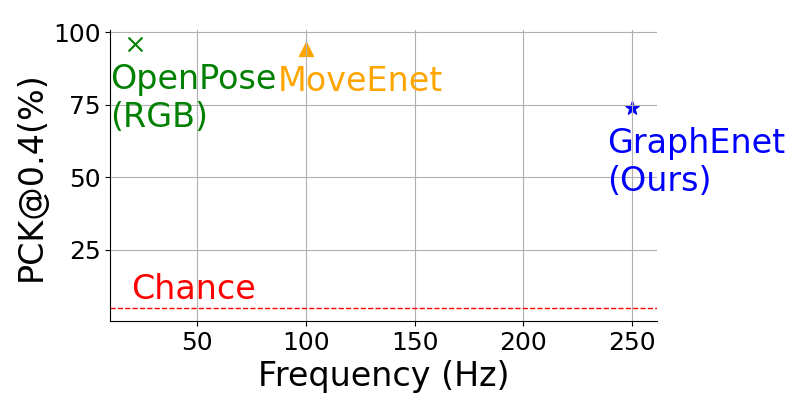}}
        \caption{The proposed GraphEnet is an initial investigation into using GNNs for human pose estimation with event cameras, with 2.5$\times$ faster than the previous state-of-the-art with a minor performance drop. The chance baseline is calculated assuming a random pixel in the image is chosen for each joint position.}   
        \label{fig:rebuttal} 
\end{figure} 

Event based cameras are bio-inspired visual sensors, consisting of an array of pixels that capture the \textit{change} in intensity of light in the scene asynchronously. If no change occurs, no data is produced; instead lighting change results in a sequential stream of only those pixels that detected change. This gives rise to a number of interesting properties, such as low energy consumption, reduced data transmission and low latency, all of which can be exploited in a range of applications where power and computation are constrained, including mobile robots and portable electronics. Embedded algorithms require similar properties (low latency and resource computation) to be ideally positioned to bridge the gap from sensory input to higher level applications while leveraging the unique properties of event cameras. In fact, complex, higher level vision applications like HPE are scarcely explored with event cameras because accurate estimation may require more computation, with high latency, negating the advantage of the event cameras. In this work, the authors explore the trade off between accuracy and latency, pushing the limit of operational frequency at the expense of accuracy.

% Event cameras can transmit small packets of individual pixels as soon as they have been triggered, rather than waiting for the full image to be ready and transmitted. As the latency is minimised, these sensors are ideal for system that prioritise the minimization of latency, like drones or mobile robots. But to leverage this property of event cameras, the algorithms based on this data also require a low latency. 

This work proposes a Graph Neural Network (GNN) for the estimation of 2D human pose based on a stream of events. %With the rise in the popularity of event based cameras it is becoming increasingly relevant to consider being able to do all major vision applications with these sensors, in order to create single sensor holistic applications. 
Graph Neural Networks \cite{zhou2020gnnsurvey} have gained traction in recent years, due to their flexibility with data input size and sparse computation, giving them an ability to adapt to a wide variety of applications, from drug discovery to social network recommendation systems; they can be applied to any type of data that can be encoded into a graph. In contrast, convolutional neural networks typically require dense data of constant size - e.g. as contained in images produced by traditional cameras. Event streams are sparsely populated, since events are only produced where and when there has been change in intensity leading to GNNs being more appropriate for such data.    

This paper investigates the potential of a GNN to estimate human pose using event camera data, and the focus is on the trade off between accuracy and latency. Previous works using events as the input to a GNN created the graph directly from events~\cite{schaefer2022aegnn}, which can be computationally heavy due to the sheer number of data points. The graph creation step is often a bottleneck in the pipeline, which adds substantial latency to the full system. This paper proposes using a sparse line based event representation to build the initial input graph for the GNN, and investigates promising GNN processing methods to learn and extrapolate the relevant information from such an input graph. The proposed architecture obtains 74\% PCKt@0.4 accuracy on event-Human 3.6 Million dataset with an update rate of 250 Hz, as shown in Fig.~\ref{fig:rebuttal}.
%The question we seek to answer is: is it viable to use a GNN for a real time HPE application? How much accuracy can be obtained from such a system? How fast can the resulting system be? Thus, the trade off between accuracy and latency can be analysed. The various design elements of the model are described and a detailed ablation study is reported, which informed the final design choices. GraphEnet achieves 74\% PCKt@0.4 accuracy on Human 3.6 dataset with an update rate of 250 Hz. 

% 
% \todo{Is it worth defining the contributions of the work here. We already summarised the novelties in the deliverable document in terms of added GNN features. Other contributions include the ablation study and the first real-time GNN from events.}

Overall, this work has the following contributions:
\begin{itemize}
    \item Propose the use of line segment features to reduce the graph building time, and dimensionality, and later reduce the computational time in the GNN layers, leading to the first real-time GNN for event cameras. %Previously, GNNs have been built directly on graphs connecting raw events, e.g.~\cite{schaefer2022aegnn}.
    % \item Learning weighted vectors to spatially transpose node position to joint position, as well as confidence of individual node contributions to each joint. 
    % \item Augmented connectivity between nodes in the initial graph formation process.
    % \item The first real-time GNN for event cameras.
    \item An architecture with lower latency than state-of-the-art methods for Human Pose Estimation with event cameras and the first GNN applied to this scenario.
    \item An in-depth ablation study to analyse the contribution of each component to the final results.
    % \item The first GNN applied to Human Pose Estimation for event cameras, which achieves 74\% PCKt@0.4 accuracy on Human 3.6 dataset with an update rate of 250 Hz
\end{itemize}

% In this work, we explored the potential for HPE with line segment inputs. Human body shapes and silhouettes are essentially a set of lines, but the length of line segments would vary, thus an accurate silhouette would not be possible with a fixed length line segment. On the other hand, estimating body joint locations may not require accurate body shapes. Thus, the question we explore in this paper is: using line segment based representations derived from event input, how well can we estimate body joints.

%%%%%%%%%%%%%%%%%%%%%%%%%%%%%%%%%%%%%%%%%%%%%%%%%%%%%%%%%%%%%%%%%%%%%%%%%%%%%%%%
\section{RELATED WORK}
%%%%%%%%%%%%%%%%%%%%%%%%%%%%%%%%%%%%%%%%%%%%%%%%%%%%%%%%%%%%%%%%%%%%%%%%%%%%%%%%
This work sits at the intersection between Graph Neural Networks, Human Post Estimation, and Event Cameras.
\subsection{Human Pose Estimation}
In recent years, numerous studies have been published on human pose estimation using RGB data, encompassing approaches from skeletal poses and localisation of specific poses to volumetric representations of the entire body~\cite{andriluka14cvprmpii,chen2020monocular,wang2021deep,liu2022recent}.
Bottom-up artificial neural network (ANN) approaches initially detect limbs before grouping them for each human, with OpenPose~\cite{cao2019openpose} being a seminal work that is still widely used because of its ease of use and accuracy, providing a solid baseline method. Top-down ANN approaches~\cite{su2019multi, qiu2020peeking} use single-person pose estimation for each individual detected in the raw input: MoveNet is such a system that runs at high-frequency, although it did not have a published paper at the time of writing~\cite{movenet_link}.

\paragraph{Event-based methods}
In the domain of event-based vision, research on HPE is relatively limited. EventCap~\cite{xu2020eventcap} used a hybrid event and frame camera for high-frequency 3D volumetric pose. However, the accuracy relied on a backwards optimisation step between frames, making it an offline-only system. EventHPE~\cite{zou2021eventhpe} uses a greyscale image to estimate initial pose and events and optical flow to estimate pose changes. These methods are less practical as they require multiple cameras.

LiftMono-HPE~\cite{scarpellini2021lifting} employs an ANN-based system to estimate a 3D skeleton pose at 2 Hz, from a monocular event camera using torso length as a depth prior. EventPointPose~\cite{chen2022efficient} estimated 2D human pose with low latency by exploiting 3D point cloud processing techniques applied to events. The baseline method for DHP19~\cite{Calabrese2019_DHP19} used stereo event cameras to perform 3D HPE. MoveEnet~\cite{goyal2023moveenet} is the only network that operated directly on a single camera's event stream without prior information. It converts the input event stream to a time invariant event surface, from which a multi-headed CNN extracts human pose. Since then, \cite{yu2024adaptiveViT} has proposed a Vision Transformer model for HPE.

EventGAN \cite{zhu2021eventgan} presents a method to pre-train models designed for event based data in order for fast model convergence, and improving on object detection and the HPE task.

\begin{figure*}
    \centering
    \includegraphics[width=0.9\linewidth]{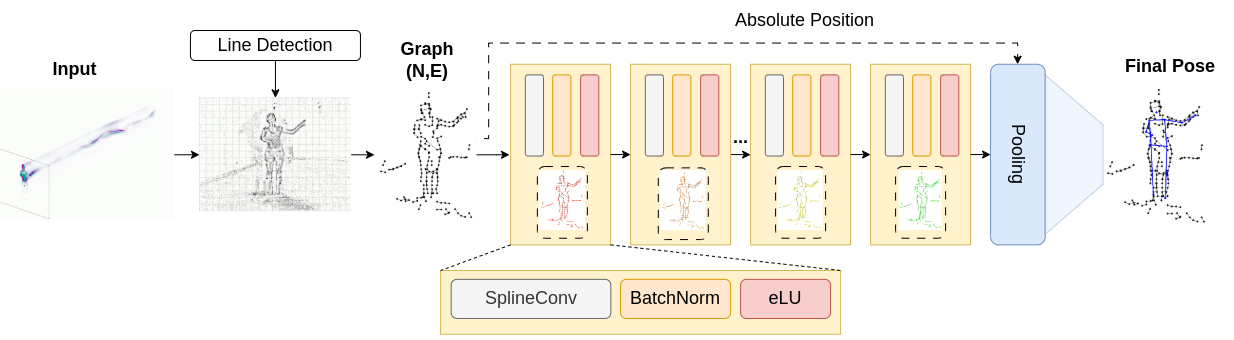}
    \caption{Pipeline of GraphEnet. The input is the continuous, asynchronous stream of events from an event camera. Line detection is performed on a velocity-invariant event accumulation, such that line segments are detected in each grid placed over the surface. The input graph is formed by connecting nearby detected line end-points. The GNN processes the graph, aggregating node features to represent each joint appearance. The final pooling layer extracts joint positions given learnt vector offsets from each node.}
    \label{fig:network-schematic} 
\end{figure*}

\subsection{Graph Neural Networks}
Graph Neural Networks (GNNs) are a class of neural networks designed to operate on graph-structured data which have found applications in vision~\cite{nazir2021surveyimageGNNs}, motion prediction~\cite{lyu20223dmotionpredictionsurvey} and human action recognition~\cite{ahmad2021GCNNforHARSurvey}. Nodes can represent superpixels (groups of pixels with a shared property) and edges are created based on region adjacency or K-nearest neighbours. There are a number of design choices: type of convolution, aggregate function, hierarchical vs flat pooling, node selection methods, number of layers and sequencing. Very few applications based on human input are explored with GNNs, most notably Human Action Recognition and motion prediction~\cite{ahmad2021graph, li2021directed}, but in most works, the human skeleton or pose is first extracted by other methods, then the GNNs are applied on the downward task. 

\paragraph{GNNs with event based cameras} 
AEGNN~\cite{schaefer2022aegnn} proposes an asynchronous event-based GNN for object recognition, creating a graph with each event is a node and which is iteratively updated as each event is produced. However, the graph building is a computationally expensive process in their system. HUGNet~\cite{dalgaty2023hugnet} proposes a lighter-weight graph building method by applying constraints to the node connection search area, performing optical flow and object recognition~\cite{sun2023objectgnn,li2021slidegcn}. EDGCN~\cite{deng2023dynamicgnn} uses distillation from frame data to the event-based graph to inform the edges for object detection and classification. \cite{gehrig2024automotivegnn} merges CNN based features from intensity images with high temporal resolution events to achieve a low latency estimation of pedestrians walking onto a busy road. 

To the best of the authors' knowledge there are no research articles applying event-based GNNs for any tasks related to human input, such as action, pose or motion estimation.

\subsection{Line segments based methods} 
Line segments have frequently been explored as an intermediate representation for vision applications with RGB cameras~\cite{lin2024linesegmentssurvey}. 
They have also been explored for a few tasks with event data. LECalib~\cite{liu2024lecalib} detects line segments in events integrated over a fixed period of time for camera calibration. Another work~\cite{chamorro2020cameratracking} uses a line-based representation to track the change in pose of a camera and later for a full SLAM pipeline~\cite{chamorro2022lineslam}. Powerline has been tracked with line estimation onboard drones fitted with event cameras~\cite{dietsche2021powerline}, by detecting planes in the spatio-temporal space of events. Other works have also proposed general methods to extract line segments from event data~\cite{wang2024evlsd,valeiras2019linefitting}.
%\todo{[TODO: add a line on how this fits.]}  %An iterative weighted least squares has been proposed~\cite{valeiras2019linefitting} fitting for each incoming event attributed to a line focusing on the line fitting itself, where the paper makes a case for its application in line-based SLAM. And EvLSD-IED~\cite{wang2024evlsd} proposes a learning based line detection method, though does not attach it to an application. Line features are shown to make a reasonable 

% \subsection{Representations}

% To take into account missing information due to non-moving parts, SCARF and line segment tracking and detection are used. 

% [ ] represented events in a binary 2D image-like matrix that indicated event presence. The surface of Active Events [ ]. 3D voxel-like structures [ ] can represent events with a third temporal dimension.  

% [Advantages of SCARF for HPE]

% [Advantages of LEDGE representation for HPE]

% \begin{figure}
%     \centering
%     \includegraphics[width=\linewidth]{presentations/media/gnn-example.png}
%     % \includegraphics[width=0.5\linewidth]{T5.8.2_images/scarf-activelayer.png}
%     \caption{A visual abstract of the work[placeholder]}
%     \label{fig:building-graph}
% \end{figure}

%%%%%%%%%%%%%%%%%%%%%%%%%%%%%%%%%%%%%%%%%%%%%%%%%%%%%%%%%%%%%%%%%%%%%%%%%%%%%%%%
\section{Methods} \label{section:methods}
%%%%%%%%%%%%%%%%%%%%%%%%%%%%%%%%%%%%%%%%%%%%%%%%%%%%%%%%%%%%%%%%%%%%%%%%%%%%%%%%

The GNN consists two main components: (1) building the initial graph from the event stream, and (2) processing the graph to aggregate information and extract most likely joint locations. An overview of the pipelin is shown in Fig.~\ref{fig:network-schematic}. %In this section we will describe each of this step for the proposed network.

\subsection{Building the graph}
\subsubsection{Event stream to line segment features}
\begin{figure}
    \centering
    \includegraphics[width=0.9\linewidth]{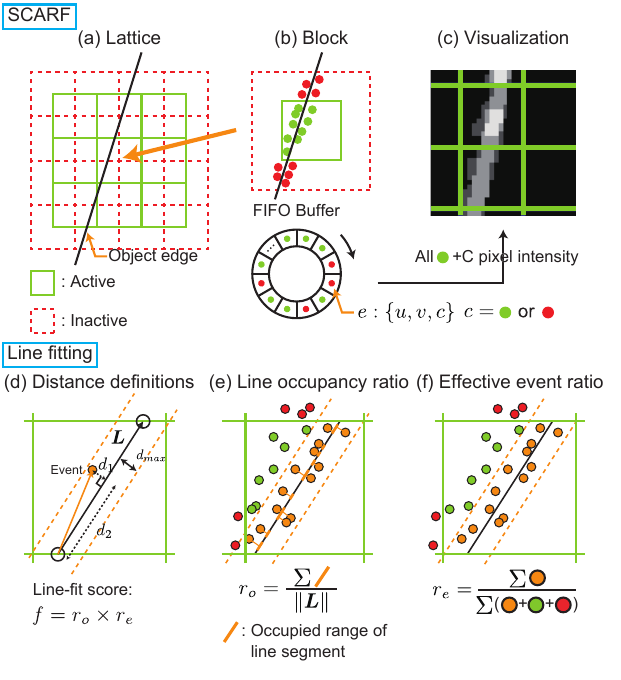}
    \caption{Algorithm overview to extract line segment features from raw events in real time \cite{ikura2025iccv-nevi}. (a) Lattice structure consists of active and (overlapping) inactive regions. (b) Each block stores active/inactive events together based on a FIFO principle. (c) Image-like representations can be generated by adding pixel intensities to all active events. (d) Line-fit score is calculated with (e) line occupancy ratio and (f) effective event ratio.}
    \label{fig:line-features-oveview} 
\end{figure}
Raw events are converted to line segment features at a very high frequency ($> 1$ kHz) using 
line detection and tracking with perturbation as in~\cite{ikura2025iccv-nevi} and overviewed in Fig.~\ref{fig:line-features-oveview}. The proposed method first uses a velocity invariant event accumulation named the Set of Centre Active Receptive Fields (SCARF), which is fast, asynchronous, and performed for each and every event. Line detection and tracking cannot be performed ``event-by-event'', and is instead run as-fast-as-possible using the most recent events in each grid-block of the SCARF representation. A single line is fit to the pixels in each block, using a random sample consensus method to fit a 2 parameter line model. Lines are rejected if there are not enough events in the block, or if the line-fit score is below a threshold.

%The events from the camera stream are accumulated into a speed invariant event surface, along the lines of~\cite{glover2024edopt}. Building the surface is done in an event-by-event manner, but is computationally light-weight such that it adds negligible latency to the pipeline (compared to GNN processing stages), even at high event rates. The surface is extracted and divided into a uniform grid. A single line is fit to the pixels in each block of the grid, using a random sample consensus method to fit a 2 parameter line model. Lines are rejected if there are not enough events in the block, or if the line-fit score is below a threshold. 

Each of the two processes, building the event surface ($\mu s$ latency) and extracting line segment features ($ms$ latency) from the event surface, can be processed simultaneously and in parallel (on a separate computational core) to the GNN components, thereby enabling a high event throughput ($>20e^{6}$ events per second) in real-time, without blocking for GNN processing. The line features can be sampled at a millisecond temporal resolution, and as required by the respective downstream steps. 

% A new graph is built whenever required by the GNN processing pipeline, using the most up-to-date event surface.
\subsubsection{Features to graph}
The group of extracted line segment features is then converted into a graph representation ($G(N,E)$ where $N$ is the set of nodes and $E$ is the set of edges), where the line segment features form the initial set $E$. A processing step is required to explicitly connect $E$ and place nodes at the connection points. The initial connection is trivial as neighboring lines have identical end-point positions. $E$ are assigned a feature describing relative position of the two nodes they connect, and $N$ are assigned features describing the pixel position on the image plane.

As there are a fixed number of blocks in the line feature method, but not each block necessarily contains a line, there is a variable number of nodes and edges in the graph. However, GNNs, differently to convolutional neural networks, are conducive to processing with variation on input size.

\subsubsection{Graph Augmentation}

\begin{figure}
    \centering
    \subcaptionbox{}[0.46\linewidth]
   {\includegraphics[width=\linewidth]{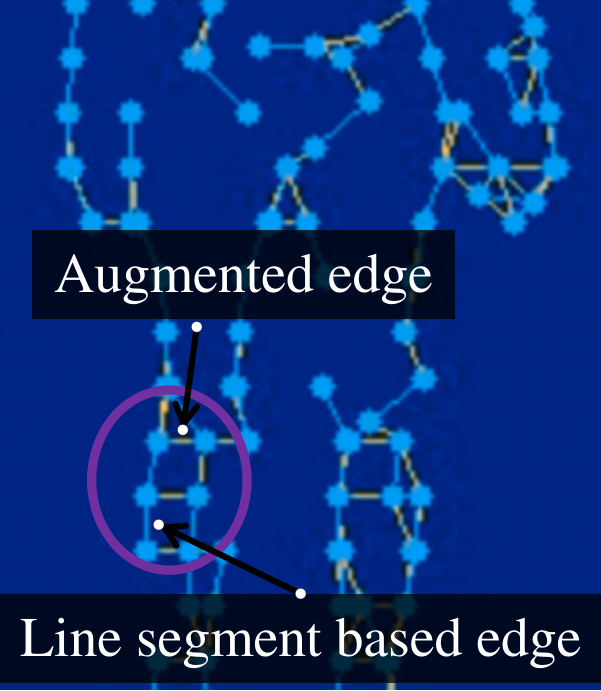}}
    \subcaptionbox{}[0.53\linewidth]{\includegraphics[width=\linewidth]{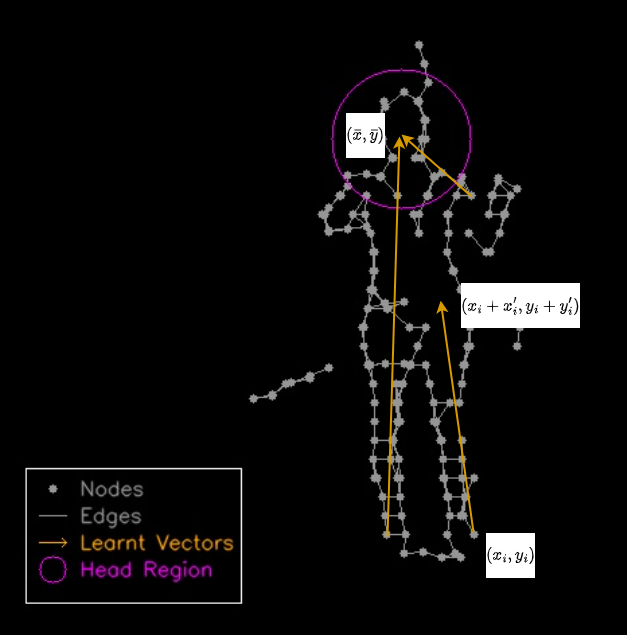}}
    \caption{Graph building and joint regression, using (a) missing edges in the graph are added if the end-points are nearby (Equation~\ref{eq:augment}), and (b) offset vectors are learned that point from node positions, to expected joint positions, given a learned confidence weighting.}\label{fig:pooling}
\end{figure}

When a graph is processed by a GNN, the features of each node are passed to neighbouring nodes connected by edges. A perfectly built graph from the input data would theoretically produce a well connected graph contouring objects in the scene, allowing each node to represent the shape of the contour at that point. Due to noise and inaccuracy, lines can be missing or edges may not align perfectly, severing the information flow through the graph. To reduce the effect of such problems, the edges in the graph have been augmented, based on the following condition: an edge is added between 2 previously unconnected nodes $N_1(x_1, y_1)$ and $N_2(x_2, y_2)$ if:
\begin{equation}\label{eq:augment}
dist(N1,N2)<\zeta,
\end{equation}%
where $dist()$ is the Euclidean distance between the points in pixels and $\zeta$ is a hyperparameter empirically set to a value slightly larger than the block size of the line segment detection step. Fig.~\ref{fig:pooling} show the difference in connectivity of a graph with augmentation applied.

% \begin{figure}[h]
%     \centering
%     \begin{tabular}{cc}

%     \subcaptionbox{$\zeta$ = 0}{\includegraphics[width=0.12\textwidth]{publications/HPE_GNN/Images/connectivity_0.png}} &
%     \subcaptionbox{$\zeta$ = 20}{\includegraphics[width=0.12\textwidth]{publications/HPE_GNN/Images/connectivity_20.png}} \\
%     \end{tabular}
%     \caption{Graph Augmentation} \vspace{-0.5cm}
%     \label{fig:connectivity_augmentation} 
% \end{figure}
% We used tracking line segments as input to our models to estimate human pose. The node features consist of the position of the end points of the line segments $(x_1, y_1)$ and $(x_2, y_2)$ in the space (u,v). Line segments also contain the time stamp $t_s$, the fitting score $e$, the status of the line segment, and the line group admin (x,y). 

\subsection{Processing the graph}

% \begin{figure*}
%     \centering
%     \includegraphics[width=0.8\linewidth]{publications/HPE_GNN/Images/SplineConv_diagram.png}
%     \caption{GraphEnet Architecture.}
%     \label{fig:netowrk-architecture} \vspace{-0.4cm}
% \end{figure*}

%The choice of network layers was crucial to the design process. 
The absolute and relative position of the nodes was strongly informative for the task, thus SplineConv layers~\cite{fey2018splinecnn} were integrated into the network as the primary message passing and aggregating layers. SplineConv function consists of a continuous spatial convolutional kernel using B-spline bases, and incorporates position data in the feature aggregation that occurs during GNN processing. As with any deep neural network, a number of SplinConv layers are stacked along with batch normalisation and non-linearity. The SplineConv kernel size is changed across the layers. Two different variation of these kernel sizes were tested: biconic (increase and then decrease, with first and last layer at similar size) and conic (gradual increase with maximum size at the last layer).

Intuitively, the process undertaken by the network layers is similar to that of a convolutional neural network. However, information is passed to the local neighbourhood of each node in a sparse manner. The receptive field of each node, or n-hop neighbourhood increases as the the graph is processed by the layers, enabling nodes to accumulate positional information of more distant nodes. Therefore, node features accumulate information that can be consistently representative of, and unique to, a specific body joint. %This should enable them to learn a representation conducive to the target task.

\subsubsection{Pooling layer}

The final layer of the network is a pooling later that clusters the nodes to obtain the 13 body joints of the human agent. To this end, we implemented a custom pooling method. 

Each node, $i$, is trained to estimate a potential location for each joint, $j$, as an offset to its own position ($x'_{i,j},y'_{i,j}$) (see Fig. \ref{fig:pooling}). Additionally, a confidence($c_{i,j}$) is associated to each offset. We tested two different variations of the confidence value: (1) a single confidence per node-joint pair, and (2) each node-joint pair has independent confidence for x and y dimensions (labeled: axis separated).

The pooled joint position $(\hat{x}_j, \hat{y}_j)$ for joint $j$ (where $j$ is removed for simplicity) given each node, $i$, is calculated by:
\begin{equation}
(\hat{x}, \hat{y}) = \left( \sum_{i=0}^{N-1} c_{x_i}(x_i + x'_i), \sum_{i=0}^{N-1} c_{y_i} (y_i + y'_i) \right),
\end{equation}
where
$\hat{x}, \hat{y}$: estimated joint position,
$N$: number of nodes,
$(x_i, y_i)$: position of the $i^{th}$ node,
$(x'_i, y'_i)$: learnt offset of the $i^{th}$ node,
and $(c_{x_i}, c_{y_i})$: confidence for the $i^{th}$ node.
For the single confidence value $(c_{y_i} = c_{x_i})$.

\subsubsection{Losses and Training}
The final model loss, $l_{total}$ incorporates a target loss and a node loss according to $l_{total} = \alpha l_{target} + \beta l_{node}$, where each loss is calculated using the mean squared error (MSE):

\begin{equation}
l_{target} = \text{MSE}((\hat{x}, \hat{y}), (\bar{x}, \bar{y})) \text{,}
\end{equation}
\begin{equation}
l_{node} = \left(\sum_{i=0}^{N-1} \text{MSE}((x_i + x'_i, y_i + y'_i), (\bar{x}, \bar{y}))\right), %\tag{4}
\end{equation}
where $(\bar{x}_j, \bar{y}_j)$ is the ground truth position of joint $j$ ($j$ removed for readability), and $\alpha$ and $\beta$ are tunable parameters.

% \[
% \phi(\omega) = 
% \begin{cases} 
% 1, & \text{if } \omega \geq c \\
% 0, & \text{if } \omega < c
% \end{cases}
% \]
Various training paradigms were tested (results reported in Tab. \ref{tab:ablation}). The basic paradigm, labelled \textit{together}, trained the full architecture with $l_{total}$ loss, with varying values of $\alpha$ and $\beta$. Paradigms \textit{node-only} and \textit{target-only} used only the $l_{node}$ ($\alpha=0,\beta=1$) and  $l_{target}$($\alpha=1,\beta=0$) respectively. Finally, the \textit{staggered} paradigm first trained only on the $l_{node}$ ($\alpha=0,\beta=1$) for 20 epochs, thus training the nodes to point to the joints. Then $l_{target}$ is included ($\alpha=1,\beta=1$) for 50 epochs, to train the the confidence values as well.

% The variations in the model are compared experimentally and the results are shown in Tab. \ref{tab:ablation}.

%%%%%%%%%%%%%%%%%%%%%%%%%%%%%%%%%%%%%%%%%%%%%%%%%%%%%%%%%%%%%%%%%%%%%%%%%%%%%%%%

\section{EXPERIMENTS} \label{section:exp}
%%%%%%%%%%%%%%%%%%%%%%%%%%%%%%%%%%%%%%%%%%%%%%%%%%%%%%%%%%%%%%%%%%%%%%%%%%%%%%%%
Experiments perform a comparison to the state-of-the-art as well as ablation studies, for both speed and accuracy.
\subsection{\textbf{Datasets}} \label{section:dataset}
The architecture was tested on two different datasets, to enable a comparison with state-of-the-art methods. Each dataset has the ground truth joint positions for HPE using motion capture systems.

The \textbf{e-Human 3.6M dataset (eH36M)} is a semi-synthetic dataset with a spatial resolution of 640x480. It is created by converting the large-scale benchmark video dataset Human 3.6 million~\cite{h36m_pami} to events (process described in~\cite{goyal2023moveenet}). The dataset consists of 7 subjects (3 female, 4 male), each performing 15 actions. 5 subjects are used in training and 2 for validation. The results shown in the paper are on the validation dataset. 

The \textbf{DHP19 dataset}~\cite{Calabrese2019_DHP19} is a large-scale event-based dataset for human pose estimation. Its 4 cameras have the resolution of 346x280, acquiring data from 17 subjects (12 female, 5 male), each performing 33 movements. The data is split by subject, 12 for training and 5 for validation.% The data is acquired from 4 cameras placed at $\pm 45^{\circ}$ and $\pm 90^{\circ}$ from the front view of the subjects.

%%%%%%%%%%%% 
\subsection{\textbf{Metrics}} \label{section:metrics}
Two standard metrics have been adopted.

\textbf{Percentage of correct keypoints (PCK)} defines a keypoint (i.e. joints) as correct if it is within some tolerance of the ground-truth position. The metric is defines as:
\begin{equation}
\frac{100}{M} \sum_{j=1}^{M} \delta(pT - d_j), \quad \text{where} \quad \delta(x) = \begin{cases} 1 & \text{if } x > 0 \\ 0 & \text{if } x < 0 \end{cases}
\end{equation}%
where $d_j$ is the Euclidean distance between the predicted and Ground Truth positions of the $j^{th}$ joint, M=13 is the number of joints/keypoints, $T$ is a threshold equal to the diagonal torso size of the human agent in pixels and $p$ is a scaling factor in the range (0,1]. Since the PCK threshold is extracted from the sample that is being evaluated, it is independent of the input resolution, the height of the subject and the distance of the subject from the camera. Using the diagonal length of the torso makes it robust to change in orientation of the camera and most camera viewpoints. %In this work, the PCK threshold is the diagonal torso length of the subject. %We use all 13 joints in the calculation of the PCK.
 
\textbf{Mean Per Joint Position Error (MPJPE)} is calculated as the average Euclidean distance between ground-truth and prediction as follows:
\begin{equation}
\text{MPJPE} = \frac{1}{M}\ \sum_{j}^{M} \|x_j - \hat{x}_j \|, 
\end{equation}%
where $N$ is the number of skeleton joints, $x_j$ and $\hat{x}_j$ are, respectively, the ground truth and predicted position of the $j^{th}$~joint  in the image space. MPJPE is frequently used in HPE benchmarks and facilitates comparison with other works, but it is affected by factors like camera resolution. 

%%%%%%%%%%%%%
\subsection{\textbf{Comparison to state-of-the-art}} \label{section:sota}

\begin{figure}
\subcaptionbox{\label{fig:results-pck-eH36M}}{\includegraphics[width= 0.9\linewidth]{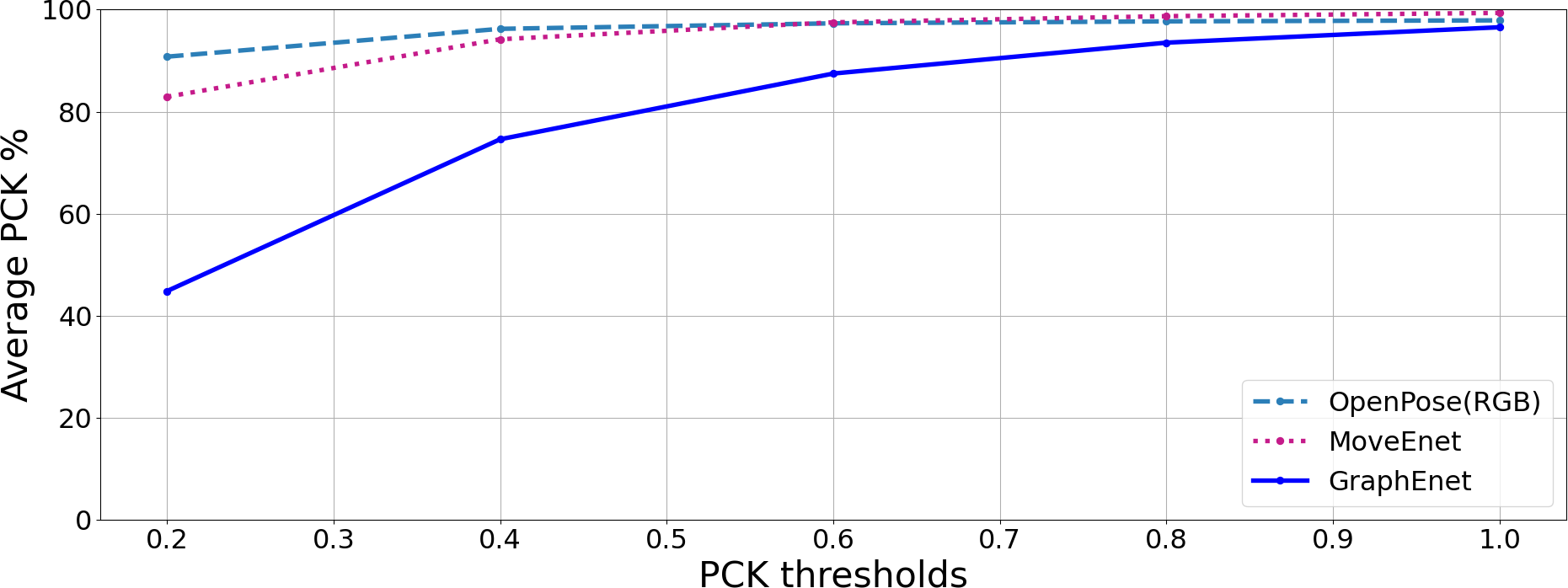}}
\subcaptionbox{\label{fig:results-pck-DHP19}}{\includegraphics[width=0.9\linewidth]{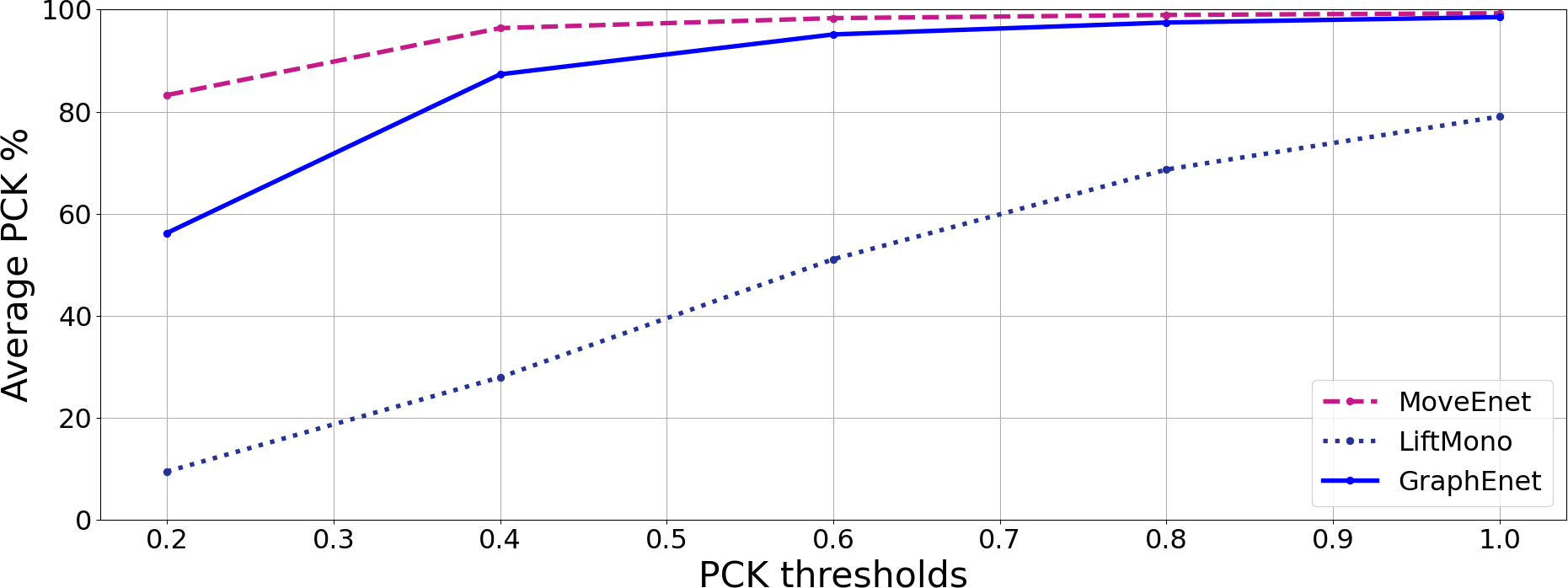} }
\subcaptionbox{\label{fig:results-pck-joints}}{\includegraphics[width=0.9\linewidth]{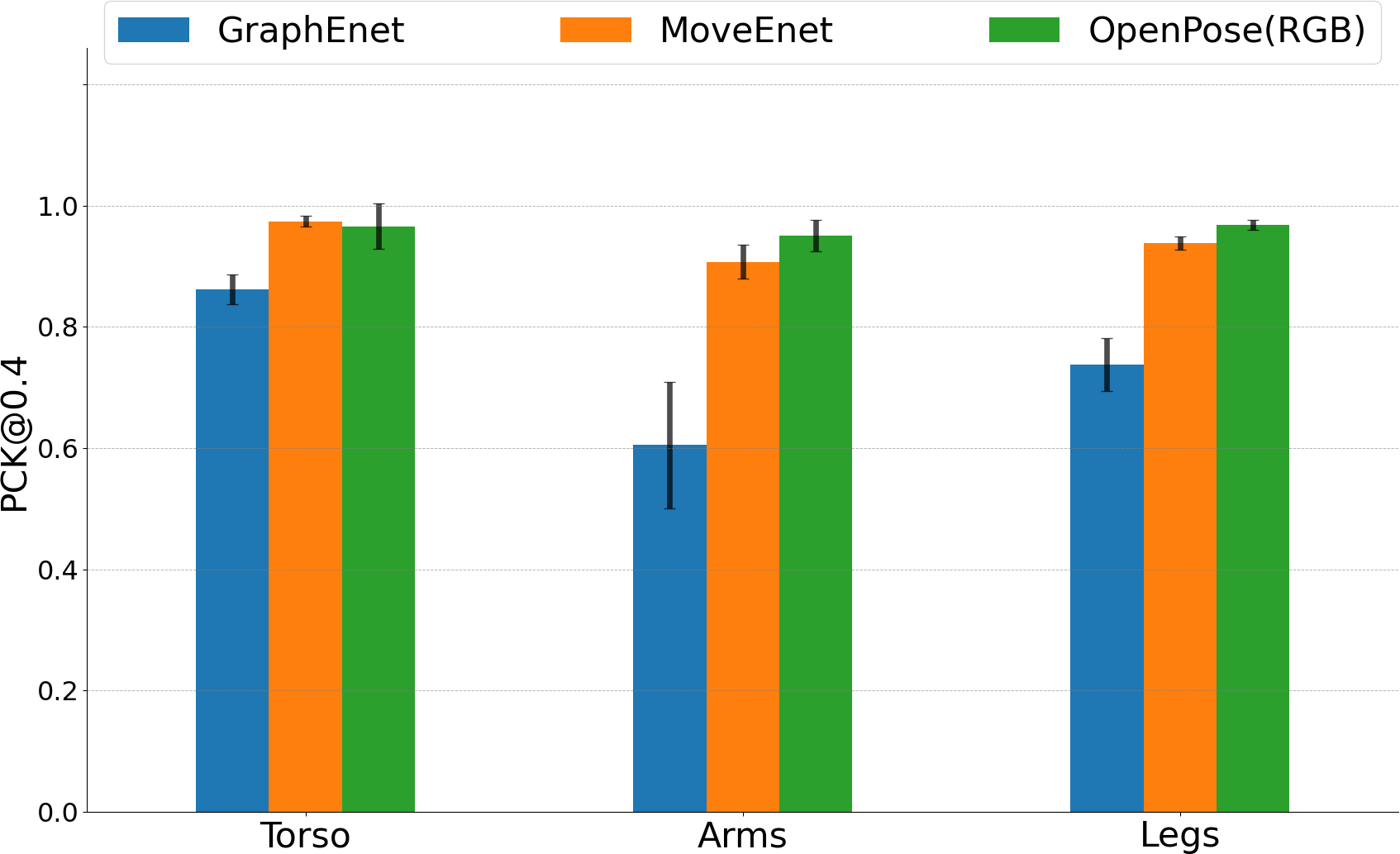}}

\caption{Comparison with state-of-the-art using a variation on PCK thresholds on the (a) eH36M dataset and (b) DHP19 dataset. The eH36M results are clustered by body-part type in (c).}\label{fig:allresults}
\end{figure}

There are few works on 2D HPE with event-cameras, and fewer still that provide results on benchmark datasets. TORE\cite{baldwin2022tore} provides results on DHP19, but separate metrics on camera 2 and 3, foregoing the other two camera viewpoints entirely. EventPointPose \cite{chen2022efficient}, an event based HPE model for 3D Pose also provides 2D results on DHP19 with the MPJPE metric and latency information. \cite{yu2024adaptiveViT} neither has results on any benchmark datasets, nor the code available at the time of writing. MoveEnet \cite{goyal2023moveenet} provides results for both DHP19 and eH36M. Results on DHP19 dataset are available from DHP19 method \cite{Calabrese2019_DHP19} (which also uses greyscale intensity values that they do not provide with the dataset) and LiftMono-HPE\cite{scarpellini2021lifting} (in \cite{goyal2023moveenet}).

Our accuracy comparison results are shown in Table~\ref{tab:global-sota-comparison} and Fig.~\ref{fig:allresults}. Since eH36M is derived from RGB data, this enables a comparison with RGB based models. The conversion is generally a lossy process that only recreates high temporal resolution through interpolation, thus an event based method on eH36M would likely not supersede the accuracy or precision of a SOTA RGB based method on the original dataset, but can provide an upper bound for context and comparison. Thus we also show results from OpenPose~\cite{cao2019openpose}.

% \begin{figure}[h]
%     \centering
%     \includegraphics[width=\linewidth]{publications/HPE_GNN/Images/qualitative_result.png}
%     \caption{Qualitative results on a sample from eH36M. 
%     a) Frame from original Human 3.6m with superimposed estimated pose of GraphEnet, MoveEnet, Openpose RGB, and GT }
%     \label{fig:results-qualitative}
% \end{figure}

% \begin{figure}
%     \centering
%     \includegraphics[width=0.9\linewidth]{publications/ICCV2025_HPE_GNN/Images/Global_MPJPE_fixed_1.png} 
%     \caption{MPJPE for different State-of-the-art models on eH36M dataset}
%     \label{fig:results-mpjpe-eH36M} 
% \end{figure}

% \begin{figure}
%     \centering
%     \includegraphics[width=0.9\linewidth]{publications/ICCV2025_HPE_GNN/Images/mpjpe_dhp19_fix_1.png} 
%     \caption{MPJPE for different State-of-the-art models on DHP19 dataset}
%     \label{fig:results-mpjpe-DHP19}
% \end{figure}

%Tab. \ref{tab:global-sota-comparison} and Fig. \ref{fig:results-pck-DHP19}  shows a comparison of GraphEnet to the aforementioned works with DHP19 dataset while Tab. \ref{tab:global-sota-comparison_h36m} and  Fig. \ref{fig:results-pck-eH36M}  show results for eH36M dataset.

\begin{table}
    \footnotesize
    \centering
    \begin{tabular}{|c|r|r|r|}
         \hline
        Model &  PCK@0.4$\uparrow$ & PCK@0.6$\uparrow$ & $MPJPE_{2D}\downarrow$  \\
        \hline
        \hline
        \multicolumn{4}{|c|}{eH36M} \\
        \hline
        OpenPose((RGB)~\cite{cao2019openpose}  & 0.96 & 0.97 & 17.07 \\
        MoveEnet~\cite{goyal2023moveenet} & 0.94 & 0.97 & 19.68 \\
        GraphEnet(Ours) & 0.74 & 0.88 & 36.21 \\
        \hline
        \hline
        \multicolumn{4}{|c|}{DHP19} \\
        \hline
        MoveEnet~\cite{goyal2023moveenet} & 0.97& 0.98 & 6.28\\
        LiftMono~\cite{scarpellini2021lifting} & 0.28 & 0.51 & 26.79 \\
        DHP19~\cite{Calabrese2019_DHP19}  & - & - & 7.03 \\
        GraphEnet(Ours) & 0.87 & 0.95 & 12.90 \\
        \hline
    \end{tabular}
    \caption{PCK comparison.}
    \label{tab:global-sota-comparison}
\end{table}

% \begin{table} 
% \footnotesize
%     \centering
%     \begin{tabular}{|c|r|r|r|} 
%     \hline
%         Model &  PCK@0.4$\uparrow$ & PCK@0.6$\uparrow$ & $MPJPE_{2D}\downarrow$  \\
%         \hline
%         MoveEnet~\cite{goyal2023moveenet} & 0.97& 0.98 & 6.28\\
%         LiftMono~\cite{scarpellini2021lifting} & 0.28 & 0.51 & 26.79 \\
%         DHP19~\cite{Calabrese2019_DHP19}  & - & - & 7.03 \\
%         GraphEnet(Ours) & 0.87 & 0.95 & 12.90 \\
%     \hline
%     \end{tabular}
%     \caption{Accuracy and Error results for DHP19 dataset}
%     \label{tab:global-sota-comparison-dhp19}
% \end{table}

% \begin{table}
% \footnotesize
%     \centering
%     \begin{tabular}{|c|r|r|r|}
%     \hline
%         Model & PCK@0.4$\uparrow$ & PCK@0.6$\uparrow$ & MPJPE$_{2D}\downarrow$ \\
%         \hline
%         OpenPose((RGB)~\cite{cao2019openpose}  & 0.96 & 0.97 & 17.07 \\
%         MoveEnet~\cite{goyal2023moveenet} & 0.94 & 0.97 & 19.68 \\
%         GraphEnet(Ours) & 0.74 & 0.88 & 36.21 \\
%     \hline
%     \end{tabular} 
%     \caption{Accuracy and Error results for eH36M dataset}
%     \label{tab:global-sota-comparison_h36m}
% \end{table}

\subsection{\textbf{Joint-wise results}}\label{section:accuracy}
 
Table~\ref{tab:joint-wisePCK} show the performance of GraphEnet on each joint in the eH36M dataset as compared to MoveEnet and OpenPose. The head is a strongly stable joint for all networks, in addition to joints on the torso. Instead, limb joints are less stable, with the line features representing them less consistently over time and body poses. To further analyse this, the 13 joints considered in this work are separated into three clusters: (1) Torso: head, shoulders and hips, (2) Legs: knees and ankles, and (3) Arms: elbows and wrists.
% \begin{enumerate}
%     \item Torso: head, shoulders and hips,
%     \item Legs: knees and ankles,
%     \item Arms: elbows and wrists
% \end{enumerate}
Fig.~\ref{fig:results-pck-joints} shows the PCK accuracy on joint clusters. GraphEnet estimates the torso joints with a higher accuracy, very close to the state of the art methods, and instead has more difficulty correctly detecting the joint positions of the legs and arms. 

%Discussion

% The accuracy of parts such as difficult joints of the elbow, wrist is further improved.

% Complexity We compare the number of parameters, computational complexity, and runtime between MoveEnet, OpenPose, and our models. 

% \begin{table}
% \centering\begin{tabular}{lrrr}
% \hline
%  Joint          &   moveEnet &   openpose\_rgb &  hpe-gnn \\
% \hline
%  head           &             0.97 &           0.93 \\
%  shoulder\_right &             0.96 &           0.96 \\
%  shoulder\_left  &             0.93 &           0.99 \\
%  elbow\_right    &             0.92 &           0.95 \\
%  elbow\_left     &             0.89 &           0.93 \\
%  hip\_left       &             0.93 &           0.94 \\
%  hip\_right      &             0.93 &           0.95 \\
%  wrist\_right    &             0.9  &           0.94 \\
%  wrist\_left     &             0.88 &           0.9  \\
%  knee\_right     &             0.93 &           0.97 \\
%  knee\_left      &             0.94 &           0.95 \\
%  ankle\_right    &             0.95 &           0.98 \\
%  ankle\_left     &             0.94 &           0.95 \\
%  avg PCK        &             0.93 &           0.95 \\
% \hline
% \end{tabular}\caption{Joint-wise PCK for threshold 0.4 on the e-human3.6m validation set}
% \label{tab:joint-wisePCK}
% \end{table}

\begin{table}
\footnotesize	
\begin{tabular}{|*{7}{c}|}
% \centering\begin{tabular}{|p{1.6cm} >{\centering}p{1.6cm} >{\centering}p{0.9cm} >{\centering}p{1.6cm} >{\centering}p{0.9cm} >{\centering}p{1.6cm} >{\centering\arraybackslash}p{2.4cm}|}
\hline
 Joint          &  \multicolumn{2}{c}{GraphEnet}   &   \multicolumn{2}{c}{MoveEnet~\cite{goyal2023moveenet}}&  \multicolumn{2}{c|}{OpenPose~\cite{cao2019openpose}}\\ \hline
Input Type      &    \multicolumn{2}{c}{Events} &  \multicolumn{2}{c}{Events} &                      \multicolumn{2}{c|}{RGB Frames}                   \\
\hline
Metric      & Accu & Error & Accu & Error & Accu & Error  \\ \hline
 head           &                                                  0.90  & 24.73 &            0.99 & 13.84 &          0.90 &  18.15  \\
 shoulder\_right &                                                  0.86 & 29.60 &           0.97 &  17.54 &        0.99 &  15.23 \\
 shoulder\_left  &                                                  0.87 &  27.82 &          0.97 &   16.19 &        0.99 &  13.00 \\
 hip\_left       &                                                  0.84 &  29.01 &          0.97 &  15.30 &        0.97 &  16.93 \\
 hip\_right      &                                                  0.84 &  29.93 &         0.97 &   17.00 &       0.98 & 20.32 \\ \hline
 elbow\_right    &                                                  0.71 &  42.35 &           0.94 & 23.03 &          0.98 & 18.02 \\
 elbow\_left     &                                                  0.68 &  39.34 &           0.92 & 20.48 &          0.96 & 15.79 \\
 wrist\_right    &                                                  0.52 &  54.73 &           0.89 & 30.15 &          0.94 & 21.01\\
 wrist\_left     &                                                  0.51 & 50.83 &            0.88 & 27.18 &          0.92 & 16.22 \\ \hline
 knee\_right     &                                                  0.79 & 33.16 &            0.95 & 18.99 &          0.98 & 14.41\\
 knee\_left      &                                                  0.77 & 32.41 &            0.94 & 17.75 &          0.97 & 15.69\\
 ankle\_right    &                                                  0.70  & 39.09 &             0.94 & 20.27 &           0.97 & 17.68 \\
 ankle\_left     &                                                  0.69 & 37.77 &           0.92 &  20.57 &         0.96 & 19.52 \\ \hline
 Average       &                                                  0.74 &   36.21 &          0.94 &   19.87 &        0.96 & 17.07 \\
\hline
\end{tabular} 
\caption{Joint-wise breakdown of the accuracy (PCKt@0.4) and error (2D MPJPE) on the eH36M dataset}
\label{tab:joint-wisePCK} 
\end{table}

\begin{table}
\small
    \centering
    \begin{tabular}{|p{3cm}|>{\centering}p{2cm}|>{\centering\arraybackslash}p{2cm}|}
    \hline
        Model & Latency $\downarrow$ & Frequency $\uparrow$ \\
        \hline
        OpenPose(RGB)~\cite{cao2019openpose} & 46ms& 20Hz \\
        MoveEnet~\cite{goyal2023moveenet} & 10ms& 100Hz \\
        GraphEnet(Ours) & \textbf{4ms}& \textbf{250Hz}  \\
    \hline
    \end{tabular}
    \caption{Table comparing with SOTA for eH36M dataset}
    \label{tab:latency}
\end{table} 
% 
% \begin{figure}
%     \centering
%     \includegraphics[width=0.9\linewidth]{publications/ICCV2025_HPE_GNN/Images/individual_wise_joints_pck_cluster_fix.png} 
%     \caption{PCK results for joint clusters, on eH36M.}
%     \label{fig:results-pck-joints} 
% \end{figure}
% 
% 
% \begin{figure}
%     \centering
%     \includegraphics[width=0.9\linewidth]{publications/ICCV2025_HPE_GNN/Images/individual_wise_joints_mpjpe_cluster_fix.png} 
%     \caption{MPJPE error for joint clusters, on eH36M.}
%     \label{fig:results-mpjpe-joints} 
% \end{figure}
% 
% %%%%%%%%%%%%%%%%%%%%%%%%%%%%%%%%%%%%%%%%%%%%%%%%%%%%%%%%%%%%%%%%%%%%%%%%%%%%%%%%%%%%%%%%%%
\subsection{\textbf{Latency}}
\label{section:latency}

% Results of computation performance

% \begin{table*}
% \footnotesize
% \centering\begin{tabular}{lcccccccc}
% \hline
% Experiment/Model & Conf. & Connect. & Layers & Feature & Training & PCK@0.4$\uparrow$ & PCK@0.6$\uparrow$ & $MPJPE_{2D}\downarrow$ \\
% \hline
% Baseline & single & 15 & 8 & Biconic & Staggared & 0.50 & 0.67 & 62.47 \\
% Dual-contribution & \textbf{axis-separated} & 15 & 8 & Biconic & Staggared & 0.45 & 0.63 & 68.02 \\
% Connectivity & single & \textbf{0} & 8 & Biconic & Staggared & 0.28 & 0.47 & 83.16 \\
% Connectivity & single & \textbf{20} & 8 & Biconic & Staggared & 0.49 & 0.63 & 67.88 \\
% % Layers & single & 15 & \textbf{6} & Biconic & Staggared & 0.45 & 0.61 & 70.48 \\
% % Layers & single & 15 & \textbf{10} & Biconic & Staggared & \textbf{0.54} & \textbf{0.70} & \textbf{58.25} \\
% Feature-shape & single & 15 & 8 & \textbf{Cone} & Staggared & \textbf{0.57} & \textbf{0.71} & \textbf{56.89} \\
% Training & single & 15 & 8 & Biconic & \textbf{node-only} & 0.21 & 0.37 & 93.01 \\
% Training & single & 15 & 8 & Biconic & \textbf{target-only} & \textbf{0.55} & \textbf{0.71} & \textbf{57.06 }\\
% Training & single & 15 & 8 & Biconic & \textbf{together} & 0.49 & 0.66 & 64.85 \\
% \hline
% \end{tabular} \vspace{-0.3cm}\caption{Ablation Study on the e-H3.6M validation set. Conf.: Confidence. Connect.: Connectivity. Feature: Shape of features over layers}
% \label{tab:ablation} \vspace{-0.3cm}
% \end{table*}

\begin{table*}
\small
\centering\begin{tabular}{lccccccc}
\hline
Experiment/Model & Conf. & $\zeta$ & Feature & Training & PCK@0.4$\uparrow$ & PCK@0.6$\uparrow$ & $MPJPE_{2D}\downarrow$ \\
\hline
Baseline & single & 15 & Biconic & Staggared & 0.50 & 0.67 & 62.47 \\
Dual-contribution & \textbf{axis-separated} & 15 & Biconic & Staggared & 0.45 & 0.63 & 68.02 \\
No Augmentation & single & \textbf{0} & Biconic & Staggared & 0.28 & 0.47 & 83.16 \\
More Augmentation & single & \textbf{20} & Biconic & Staggared & 0.49 & 0.63 & 67.88 \\
% Layers & single & 15 & \textbf{6} & Biconic & Staggared & 0.45 & 0.61 & 70.48 \\
% Layers & single & 15 & \textbf{10} & Biconic & Staggared & \textbf{0.54} & \textbf{0.70} & \textbf{58.25} \\
Feature-shape & single & 15  & \textbf{Cone} & Staggared & \textbf{0.57} & \textbf{0.71} & \textbf{56.89} \\
Training & single & 15  & Biconic & \textbf{node-only} & 0.21 & 0.37 & 93.01 \\
Training & single & 15  & Biconic & \textbf{target-only} & \textbf{0.55} & \textbf{0.71} & \textbf{57.06 }\\
Training & single & 15  & Biconic & \textbf{together} & 0.49 & 0.66 & 64.85 \\
\hline
\end{tabular} 
\caption{Ablation Study on the e-H3.6M validation set. Conf.: Confidence. $\zeta$: Threshold for additional augmented edges. Feature: Shape of features over layers} 
\label{tab:ablation}%\vspace{-0.2cm}
\end{table*}

Appropriate algorithms can leverage the low latency of event cameras without adding excessive latency themselves, in order to create a high frequency pipeline. With the parallelisation of processes, the frequency of the full pipeline can be defined by the slowest thread. Table~\ref{tab:latency} reports the processing time and effective latency of the slowest process for each algorithm. Components of the pipeline are considered different modules if they can be parallelised and the reported latency is the maximum of such a module. Tests for OpenPose~\cite{cao2019openpose} and MoveEnet~\cite{goyal2023moveenet} are done on a an Intel Core i9-10980HK @2.40GHz x 16 and NVIDIA GeForce RTX 2070 while GraphEnet was tested on an Intel(R) Core(TM) i7-9750H CPU @ 2.60GHz x 12 with a NVIDIA GeForce GTX 1650. The slight difference in hardware was due to implementation constraints, but it is expected that GraphEnet will actually run faster if tested using the more powerful i9 machine.

GraphEnet is the first event-based GNN that can perform the target task in real time, and the pipeline can be parallelised to obtain a runtime frequency of 250Hz. The trade off between the accuracy and latency of the various methods is visualised in Fig. \ref{fig:rebuttal}.

\subsection{\textbf{Ablation studies}}

Tab.~\ref{tab:ablation} reports the various network architectures and their resulting accuracies. Each model was trained on a subset of eH36M for 50 epochs. 

%Considering the various components in this method, a detailed ablation study has been conducted to promote a better understanding of the underlying mechanisms of the model. Tab. \ref{tab:ablation} describes the various components, their variations and the resulting PCK and MPJPE results. 

Since a large amount of training was necessary to produce the ablation comparison, for smarter use of resources, the training was conducted on a defined fraction of the training split eH36M dataset.
%, while the results are instead using the full validation split. 
Based on the observations of the ablations, the features indicating highest accuracy include: single confidence, augmentation threshold of 15, cone shaped node features, with target-only training. The above system was trained on the full training split, and used for results in Section~\ref{section:accuracy} and Section~\ref{section:latency}.

\begin{figure*}
    \centering
    % \begin{tabular}{llllrrrr}

    {\includegraphics[width=0.12\textwidth]{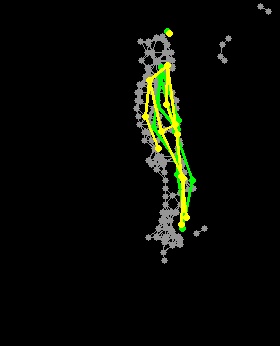}} 
    {\includegraphics[width=0.12\textwidth]{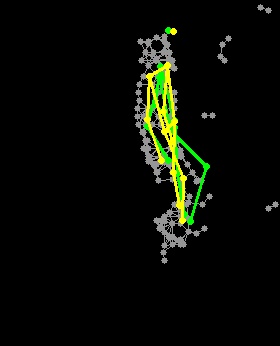}}    {\includegraphics[width=0.12\textwidth]{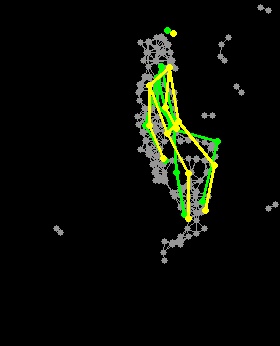}} 
    {\includegraphics[width=0.12\textwidth]{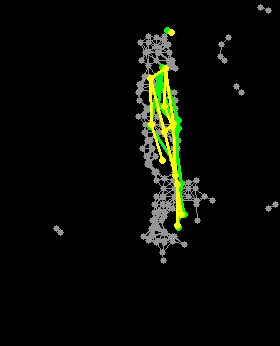}}     {\includegraphics[width=0.12\textwidth]{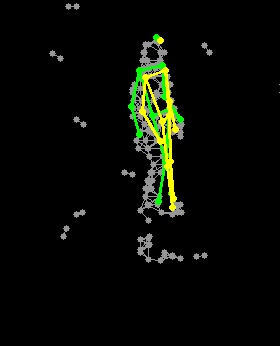}}     {\includegraphics[width=0.12\textwidth]{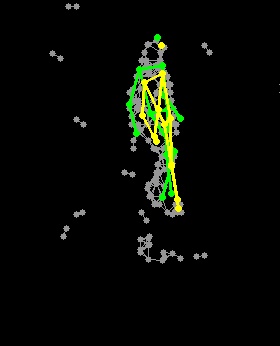}}     {\includegraphics[width=0.12\textwidth]{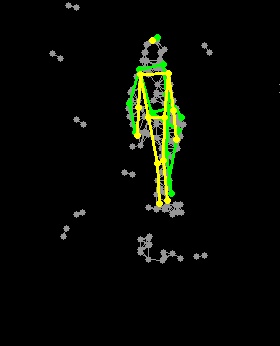}} 
    {\includegraphics[width=0.12\textwidth]{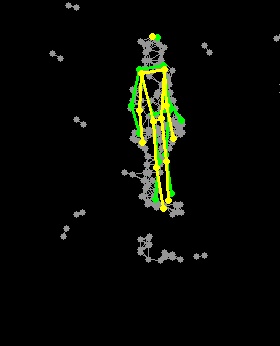}} \\

    % \end{tabular}
    \caption{Qualitative results on a sample from DHP19. Ground Truth in green, GraphEnet in yellow}
    \label{fig:qualitative_results_dhp19}
\end{figure*}

\begin{figure*}
    \centering
    \begin{tabular}{cccc}

    {\includegraphics[width=0.18\textwidth]{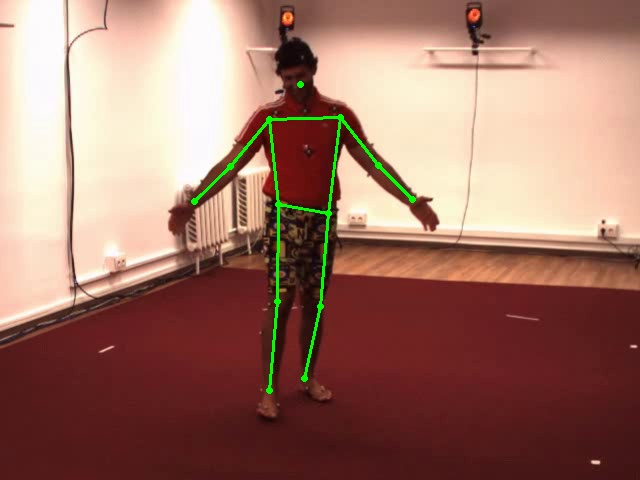}} &
    {\includegraphics[width=0.18\textwidth]{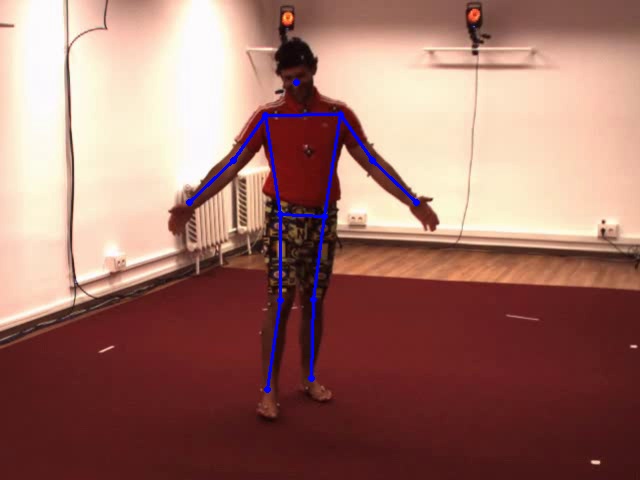}} &    
    {\includegraphics[width=0.18\textwidth]{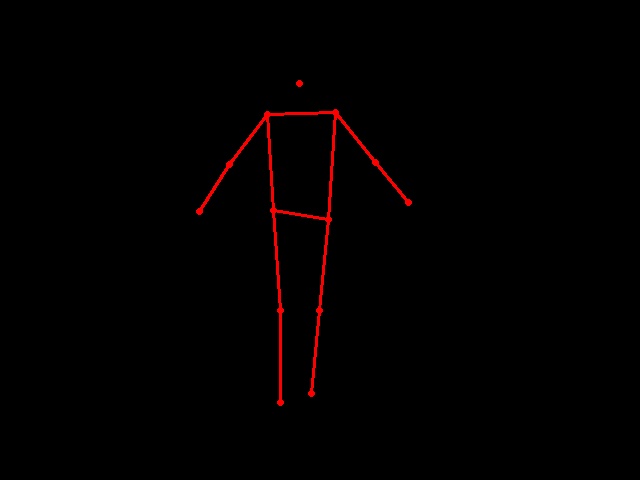}} &
    {\includegraphics[width=0.18\textwidth]{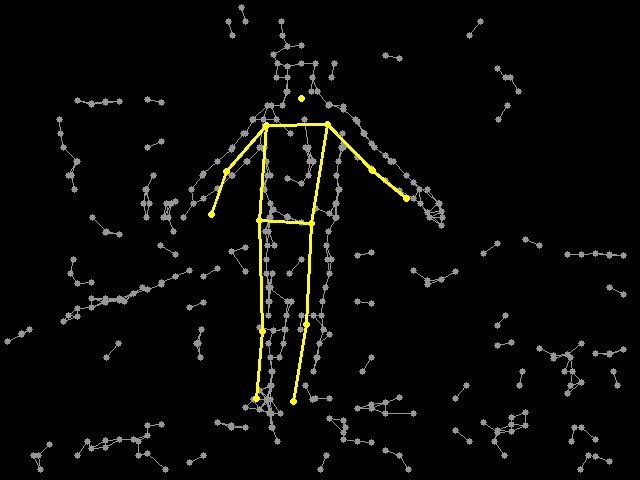}} \\
    {\includegraphics[width=0.18\textwidth]{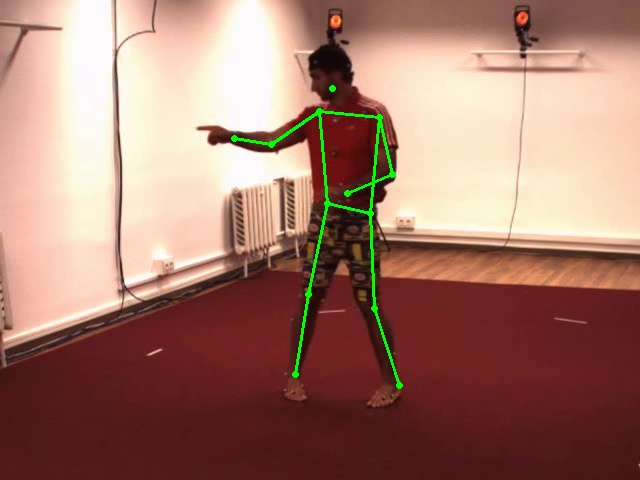}} &
    {\includegraphics[width=0.18\textwidth]{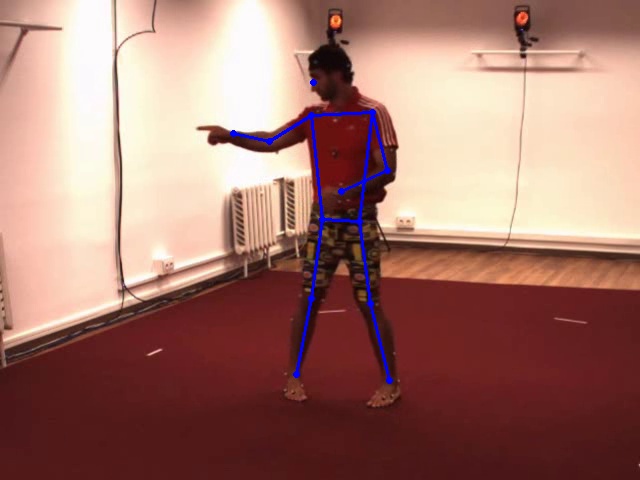}} &    {\includegraphics[width=0.18\textwidth]{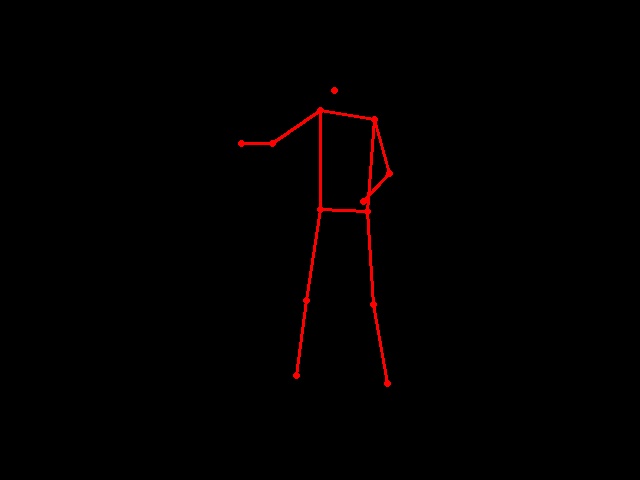}} &
    {\includegraphics[width=0.18\textwidth]{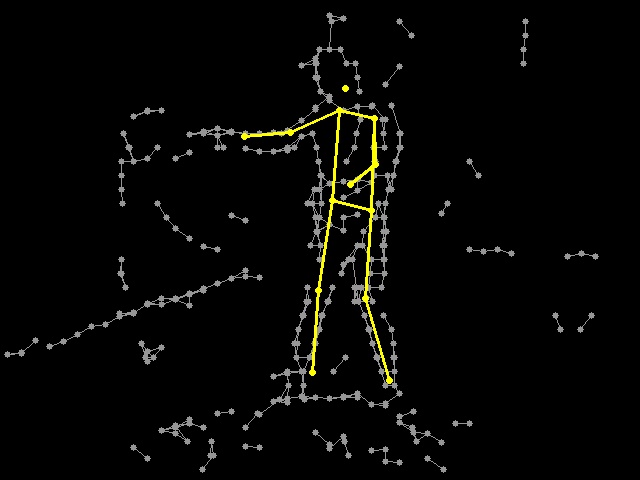}} \\
    % \subcaptionbox{Ground Truth}
    {\includegraphics[width=0.18\textwidth]{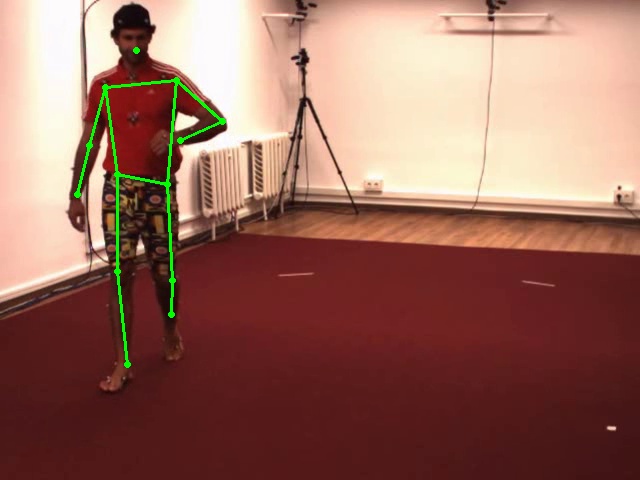}} &
    % \subcaptionbox{OpenposeRGB}
    {\includegraphics[width=0.18\textwidth]{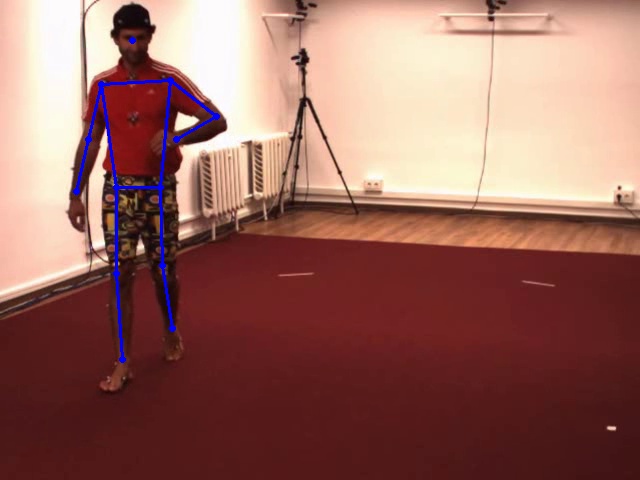}} &    
    % \subcaptionbox{MoveEnet}
    {\includegraphics[width=0.18\textwidth]{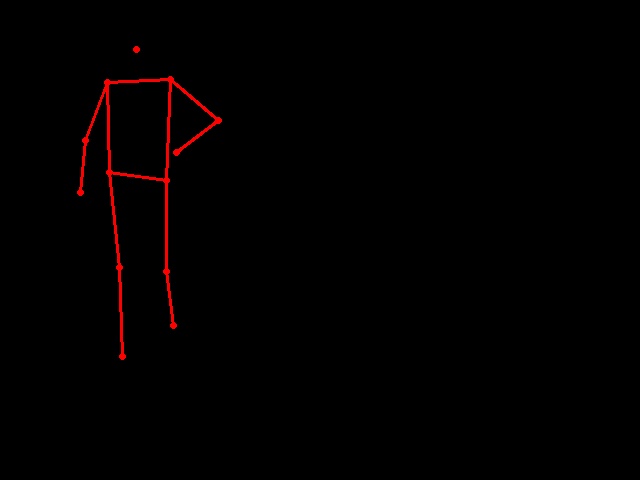}} &
    % \subcaptionbox{GraphEnet}
    {\includegraphics[width=0.18\textwidth]{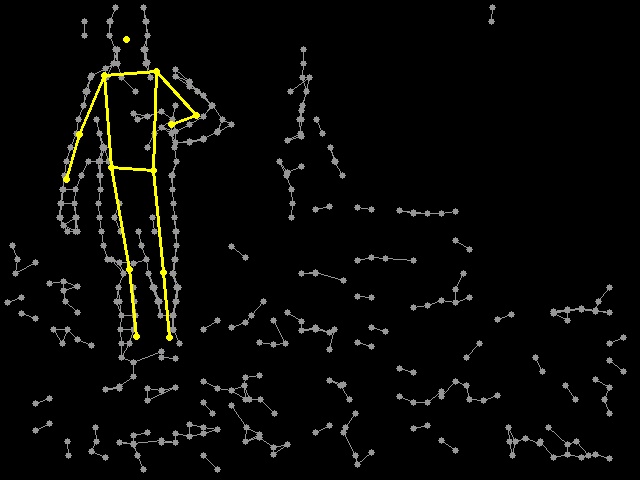}} \\

    \subcaptionbox{Ground Truth}{\includegraphics[width=0.18\textwidth]{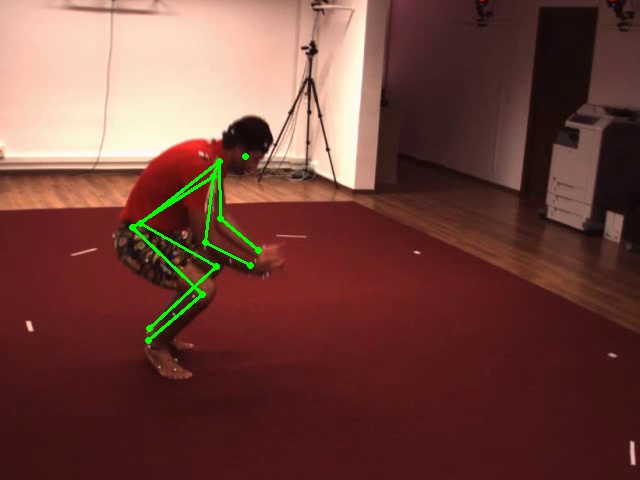}} &
    \subcaptionbox{Openpose(RGB)~\cite{cao2019openpose}}{\includegraphics[width=0.18\textwidth]{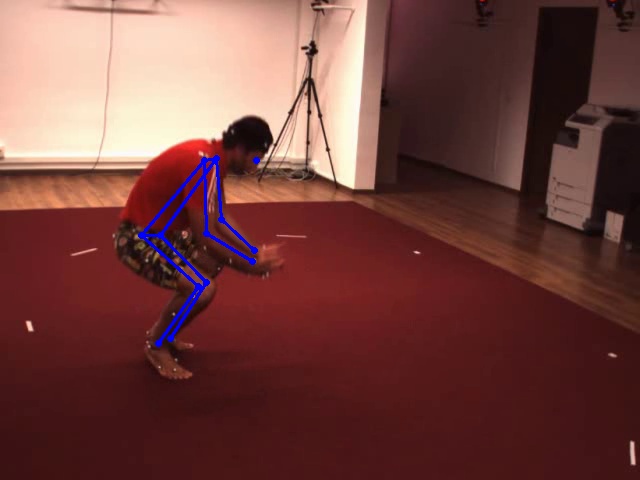}} &    \subcaptionbox{MoveEnet~\cite{goyal2023moveenet}}{\includegraphics[width=0.18\textwidth]{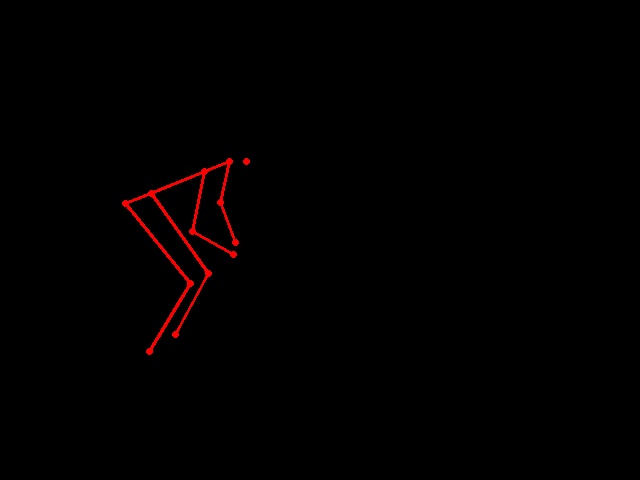}} &
    \subcaptionbox{GraphEnet(Ours)}{\includegraphics[width=0.18\textwidth]{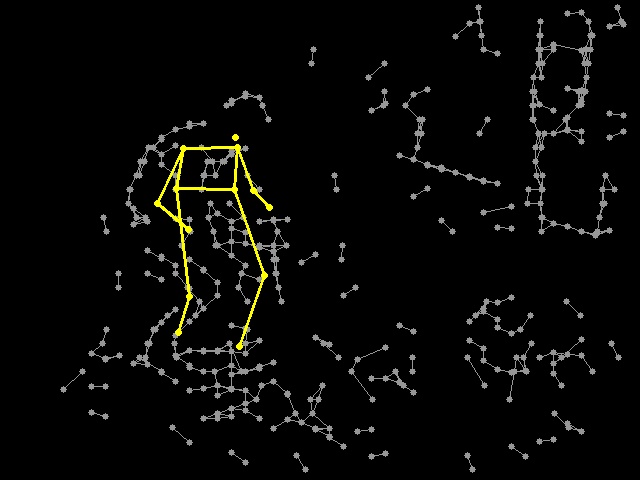}}
    \end{tabular} 
    \caption{Qualitative results on samples from eH36M}
    \label{fig:qualitative_results_h36m}
\end{figure*} 

\subsection{\textbf{Qualitative Results}}
Fig.~\ref{fig:qualitative_results_dhp19} and \ref{fig:qualitative_results_h36m} show the qualitative results of GraphEnet from DHP19 and eH36M, respectively. The prediction on DHP19 superimposes on the ground truth substantially, as is expected from the quantitative results. Results are consistent even with self occlusion. Fig. \ref{fig:qualitative_results_h36m} shows that the model performs reasonably well even where there are noisy line segments present in the scene (see row 3) but tends to fail when the noise superimposes on the person (see row 4) or the information from the joints is too sparse. The level of granularity and detail in the extremities (hands and feet) varies in the graph, thus leading to a lower robustness in the estimation of resulting pose, consistent with the quantitative results.

% \todo{write some more about what is seen here. maybe that the first 3 rows show good results, while GraphEnet shows difficulty in precise joint position for the bottom two rows}

%%%%%%%%%%%%%%%%%%%%%%%%%%%%%%%%%%%%%%%%%%%%%%%%%%%%%%%%%%%%%%%%%%%%%%%%%%%%%%%%
\section{DISCUSSION} \label{section:discuss}
%%%%%%%%%%%%%%%%%%%%%%%%%%%%%%%%%%%%%%%%%%%%%%%%%%%%%%%%%%%%%%%%%%%%%%%%%%%%%%%%
GraphEnet performs well on DHP19 samples and shows promising results on eH36M, but with a lower latency and higher frequency compared to other state-of-the-art methods.
 
GraphEnet can distinguish the body from the noisy background based on the line connectivity. The torso joints are estimated well and in a stable manner. However, arms and legs are more challenging for the GNN to detect correctly. The input graphs show that they have less consistent input information about the limbs. The area occupied by the limbs is lower compared to the torso, leading to less nodes, and the intricacy of the joints (e.g. hands and fingers) is not well represented by line segments. The low spatial resolution of the two datasets is also detrimental to finer details. 
 
Finding a balance between speed (i.e. graphs from line segments) and detail (i.e. graphs directly from events) is needed in the graph-making process. The line segments features build a graph, lowering latency, but a hierarchical model can be developed to obtain more detailed graph edges around the regions of the hands to improve accuracy at a relatively small cost of latency, in the future. The novelties proposed in the paper made it possible to achieve the target task. The ablation study indicates that the model with a single confidence, with a connectivity slightly larger than the block size, 10 layers, conic features size and trained on the target loss performs the best. Pretraining maybe also mitigate these issues with methods like \cite{zhu2021eventgan} or \cite{carissimi2022eros-like_mpii}.

While the proposed model is promising, there is scope for improvement. Heterogeneous graphs can be adopted, bridging the gap between line segment features and events, creating more detailed nodes where needed. An asynchronous approach can also be adopted, updating the input graph when there is change in the scene, and updating only relevant parts of the GNN (like in \cite{schaefer2022aegnn}).

% \begin{itemize}
%      \item The overall precision of joints is currently lower with GraphEnet. However, the frequency and latency is much higher. The potential for GNN for event-based visual processing is evident.
%      \item The  
%      \item 
%      \item The novelties introduced by the network are necessary to have the GNN work (although we don't show a vanilla GNN).
%      \item The ablation study indicates that the X model is the best - why might that be?
%      \item iterative graph building could be adopted for this system as well?
%  % \end{itemize}

%%%%%%%%%%%%%%%%%%%%%%%%%%%%%%%%%%%%%%%%%%%%%%%%%%%%%%%%%%%%%%%%%%%%%%%%%%%%%%%%
\section{CONCLUSION} \label{section:conc}
%%%%%%%%%%%%%%%%%%%%%%%%%%%%%%%%%%%%%%%%%%%%%%%%%%%%%%%%%%%%%%%%%%%%%%%%%%%%%%%%
We introduced GraphEnet for human pose estimation with a graph neural network. GraphEnet achieved reasonable (74\% v.s. 94\%) accuracy performance compared to the state-of-the-art, but with a lower latency (4 ms v.s. 10 m.s.) and higher output frequency(250 Hz v.s. 100 Hz). These preliminary results suggest that the use of GNNs can reduce latency and computational cost, while increasing the estimation frequency. 

The confidence-based offset learning in the final pooling layer of the proposed architecture was necessary for strong performance. Such a pooling layer would also be useful for other tasks, such as object recognition. The comprehensive ablation study further contributed to understanding of the individual impact of the various architectural and training components. 

GNNs, together with event cameras, enable fully sparse data pipelines realising an overall lower computational cost. Such systems can reduce the energy cost of AI systems and reduce the amount of GPU servers required, leading to positive impacts on the environment.

\section*{ACKNOWLEDGEMENTS}

This work was funded and supported scientifically by Sony Interactive Entertainment.

{
    \small
    \bibliographystyle{ieeenat_fullname}
    %\addtolength{\textheight}{-4cm}
    \bibliography{bibliography.bib}
}

%%%%%%%%%%%%%%%%%%%%%%%%%%%%%%%%%%%%%%%%%%%%%%%%%%%%%%%%%%%%%%%%%%%%%%%%%%%%%%%%

\end{document}